# Performance of large language models in the optical diagnosis of colorectal polyps

Joshua C. Vences*[1]; William T. Tran*, MSc[1]; Nikko Gimpaya, MEd[1]; Catharine M. Walsh, MD, MEd, PhD[2, 3]; Rishad J. Khan, MD[4]; Robert Bechara, MD, FRCPC[5]; Asher C. Wiggins[1]; Celine N. Rousan[1]; Kaitlyn V.G.L. Morgado[1]; Angie Ibrahim[1]; Kevin H. M. Kuo MD, MSc[1,6,7]; Daniel von Renteln, MD[8,9]; Alexander Hann, MD[10], Dennis L. Shung, MD, MHS, PhD[11]; Michael A. Scaffidi, MD, MEd[1], Charles Ménard, MD[12,13]; Joshua Landy, MD[1,6]; Samir C. Grover, MD, MEd, AGAF[1,6,14]

1. Scarborough Health Network Research Institute, Toronto, Ontario, Canada
2. Division of Gastroenterology, Hepatology and Nutrition, and The SickKids Learning and Research Institutes, The Hospital for Sick Children, Toronto, Ontario, Canada
3. Department of Paediatrics, Temerty Faculty of Medicine, University of Toronto, Toronto, Ontario, Canada
4. Division of Gastroenterology, University of Calgary
5. Division of Gastroenterology, Department of Medicine, Queen's University and Kingston Health Sciences Centre, Kingston, Ontario, Canada
6. Department of Medicine, University of Toronto, Toronto, Ontario, Canada
7. Scarborough Health Network, Division of Hematology, Toronto, Ontario, Canada
8. Université de Montréal, Department of Medicine, Montreal, Quebec, Canada
9. Centre de Recherche du Centre hospitalier de l'Université de Montréal (CRCHUM), Montreal, Quebec, Canada
10. Interventional and Experimental Endoscopy (InExEn), Department of Internal Medicine II, University Hospital Würzburg, Würzburg, Germany
11. Division of Gastroenterology and Hepatology, Mayo Clinic, Rochester, Minnesota, United States
12. Université de Sherbrooke, Faculty of Medicine and Health Sciences, Sherbrooke, Quebec, Canada
13. Centre Hospitalier Universitaire de Sherbrooke, Division of Gastroenterology, Sherbrooke, Quebec, Canada
14. Division of Gastroenterology and Hepatology, University of Toronto, Toronto, Ontario, Canada

* Contributed equally

Grant support: This study was funded by the Pialis Family Chair in Education, held by SCG. The funder had no role in study design, data collection, analysis, interpretation, or the decision to submit the manuscript.

Corresponding author:
Samir C. Grover, MD, MEd, FRCPC, AGAF
Scarborough Health Network Research Institute
2867 Ellesmere Road
Scarborough, Ontario M1E 4B9
sgrover@shn.ca
https://orcid.org/0000-0003-3392-1220

Disclosures: JCV, WTT, NG, CMW, RJK, RB, ACW, CNR, KVGLM, AI, KHMK, DvR, AH, DLS, MAS, CM, and JL have no conflicts of interest. SCG reports equity in and ownership of Volo Healthcare and honoraria from Sanofi.

Author contributions: JCV, WTT, NG, JL, and SCG conceived and designed the study. SCG, CMW, RJK, and RB acquired the data. JCV, WTT, NG, ACW, CNR, KVGLM, and AI analyzed and interpreted the data. JCV, WTT, and NG drafted the manuscript. All authors critically revised the manuscript for important intellectual content. NG provided statistical expertise. SCG and JL obtained funding. SCG and JL supervised the study. All authors approved the final version to be published and agree to be accountable for all aspects of the work.

Data transparency statement: Deidentified model outputs and analysis code will be available on GitHub upon publication.

**Word Count: 2837 (Body) 228 (Abstract) 190 (Lay Abstract)**

**Abstract**

**Background and Study Aims:** Accurate optical diagnosis of colorectal polyps guides resection strategy and surveillance, with multimodal large language models (MLLMs) showing potential for image-based diagnosis. We aimed to evaluate the diagnostic accuracy of MLLMs in classifying colorectal polyps and predicting histology.

**Methods:** We conducted a retrospective diagnostic performance study using the PRIME dataset, a curated set of white light and narrow-band imaging (NBI) images. We evaluated Claude Opus 4, Google Gemini 2.5 Pro, GPT-o3, GPT-4o, and GPT-5. For Paris, Narrow-band Imaging Colorectal Endoscopic (NICE), and predicted histology, we calculated F1 scores, percent correct scores, and accuracy of each MLLM compared to expert responses for 132 cases. Cochran's Q and McNemar's Test were used to determine differences between predicted values of each MLLM.

**Results:** The F1 scores among MLLMs were >0.9 for all models for neoplastic vs. non-neoplastic polyps. Gemini 2.5 Pro demonstrated the highest F1 scores for invasive vs. non-invasive polyps and low- vs. high-grade adenoma, at 0.560 and 0.492 respectively. Claude Opus 4 and GPT-5 had statistically significantly higher percent correct scores than other MLLMs at 41.7%, using Paris classification.

**Conclusions:** Claude Opus 4 and Gemini 2.5 Pro showed the highest accuracy in differentiating polyp subtypes, performing closest to expert consensus. Sensitivity and specificity, however, did not meet ESGE standards, highlighting the need for prospective multicenter trials and the design of human-in-the-loop workflows before clinical deployment.

**Lay Abstract**

During a colonoscopy, doctors examine small growths called polyps in the bowel. They must decide on the spot whether a polyp is harmless or could one day turn into cancer. This choice shapes whether the polyp is removed and when the patient returns for another test. Judging polyps from images is hard, and many doctors have had little training in it.

We tested whether advanced artificial intelligence programs, the kind behind ChatGPT and other popular chatbots, could judge polyps the way an expert would. We showed five programs images of 132 polyps and compared their answers with expert diagnoses.

The best programs correctly identified whether a polyp could become cancer more than 80 percent of the time. This came close to, but did not fully meet, the standards expert groups recommend. We also found that how we worded our instructions for the programs to follow changed how well the programs performed.

Unlike older computer tools, these programs can explain their thinking in plain words. This makes them promising for training new doctors and, in time, for helping plan follow-up care. More testing is needed before they are used in clinics.

## Introduction

Colonoscopy prevents colorectal cancer through the detection and removal of precancerous polyps [1, 2]. During colonoscopy, high quality optical diagnosis supports real-time decision making, which includes selecting a resection technique, anticipating the need for advanced resection, and setting surveillance intervals. To assist in this, endoscopists rely on validated visual classification systems, including the Paris and NICE classifications, to translate morphology, surface, and vascular patterns into categories [3-5].

Despite the availability of structured criteria, however, performance varies across endoscopists. Studies on endoscopic experts demonstrate that they are inadequately trained in optical diagnosis, demonstrating poor performance in T1 colorectal carcinomas [6]. Furthermore, learners often struggle to apply visual rules consistently, which contributes to inter-observer variability and uneven adoption in routine care [7-9]. For this reason, real-time artificial intelligence (AI) including computer-aided detection (CADe) and computer-aided diagnosis (CADx) can assist endoscopists in decision-making in polyp management [10]. When practitioners apply optical diagnosis reliably, resect-and-discard and diagnose-and-leave strategies can reduce unnecessary pathology use and procedural risks, which lowers cost without sacrificing safety under specified thresholds [3, 8, 11].

In other areas of imaging in medicine, traditional computer-aided systems improved detection in several trials, but improvements in classification performance often failed to generalize across institutions, devices, and imaging settings [12]. In parallel, vision-capable MLLMs entered clinical evaluation. Early appraisals reported both opportunities and pitfalls for GPT-4 in medical imaging [13-15]. Across specialties, MLLMs showed mixed performance in radiology tasks, while ophthalmology benchmarks showed better recognition when images are supplemented with text [13, 14]. In histopathology, in-context learning and structured prompting shifted general-purpose MLLMs toward domain behavior without supervised retraining [15]. Furthermore, broad radiology overviews outline the rapid evolution of MLLMs with reasoning capabilities and emphasize the need for robust validation before clinical use [16, 17]. In gastrointestinal endoscopy, GPT-4o and Gemini 1.5 Pro have demonstrated ability to detect polyps at a level similar to traditional CADe systems [18].

In this study we compared multiple MLLMs on curated, expert-assessed polyp images. We measured accuracy and agreement for Paris and NICE classifications and predicted histology.

Our aim was to determine the current sensitivity, specificity, and diagnostic accuracy of modern MLLMs and compare them to CADx models, human endoscopists, and societal consensus benchmarks.

## METHODS

This was a retrospective cohort study that explored the diagnostic performance of commercially available MLLMs. We used polyp images curated for the Polyp Recognition Improvement Module for Endoscopists (PRIME) Study [19]. The dataset consisted of images of the same polyp, captured using white light endoscopy (WLE) and narrow-band imaging (NBI) in three Canadian hospitals. This dataset was used to measure the improvement of endoscopists' optical diagnostic capabilities through the use of a virtual simulation platform. A gold standard was established via independent optical diagnosis by a committee of two expert endoscopists using the Paris and NICE categories [4-5]. Polyp histology was then predicted. A third expert rectified disagreements. The incidence of each polyp classification can be found in **Supplementary Table 1**.

We followed the STARD 2015 guidelines, which outline best practices for diagnostic accuracy studies [20].

Ethics approval for the retrospective analysis of the PRIME dataset was obtained from the Scarborough Health Network Research Ethics Board (REB# GAS-25-012).

### Multimodal Large Language Models

We evaluated each case using five multimodal MLLMs including four reasoning focused models (Claude Opus 4, Google Gemini 2.5 Pro, GPT-o3, GPT-5) and a generalist model (GPT-4o). These models were chosen for their high benchmark scores, common usage, multimodal capabilities, and availability via stable APIs, specifically OpenAI's API platform and Google's Vertex AI platform [21, 22]. They were also chosen to expand upon the imaging literature, as studies primarily use OpenAI's GPT models [12, 13-15]. For each model we recorded the provider, exact model ID, temperature, and access dates.

There is uncertainty in how to engineer MLLM prompts for medical imaging tasks. Longer prompts have been shown to generally improve accuracy, but ideal prompt format varies by domain [23, 24]. To investigate this, we designed two prompt modalities: one using a structured pattern that is common in clinical and imaging applications (long prompt) and one that includes a brief role and output schema (short prompt). The purpose of this was to test whether long, structured prompts improve the percentage of correct answers given the advent of MLLMs with inbuilt chain-of-thought (CoT) reasoning. The long, structured pattern includes role and objective specification, warnings, a definitions primer, a clinical decision-making algorithm, and a strict JSON output format with preset values, in concordance with the PRIME dataset, for the MLLM to select from. WLE and NBI image pairs were presented to the MLLMs within the same prompt. Prompts were approved by consensus of a group of investigators (JCV, WTT, NG, JL, SCG). A copy of both prompts appears in **Supplementary Methods 1**.

All images were de-identified at source. We removed any overlaid text and resized images to center polyps. Text and metadata were withheld from the MLLMs. No size or location information was provided to the models. Each case was evaluated ten times per model with an identical prompt and fixed parameters. We used the most frequent diagnostic classification from the MLLMs across replicates as the primary predicted classification.

**Outcome Measures and Data analysis**

The **F1 score** was selected as the primary performance metric for model evaluation, given the unequal distribution of lesion classes in the study dataset and the potential for overall accuracy to overestimate performance in imbalanced settings. Because F1 does not have a known sampling distribution, statistical comparisons between models were conducted using paired accuracy proportions. Secondary outcome measures included sensitivity, specificity, positive predictive value (PPV), and negative predictive value (NPV). Secondary analyses compared performance between long and short prompt modalities and model performance in classifying polypoid versus non-polypoid lesions.

To explore differences in diagnostic performance, the MLLM responses and gold standard were coded into binary categories (neoplastic vs non-neoplastic, invasive vs. non-invasive, and high grade adenoma vs. low grade adenoma) across classification systems (Paris, NICE) and predicted histology [3, 25]. Additional analyses were performed for polypoid vs. non-polypoid lesions using the Paris classification system. For each binary task, sensitivity, specificity, PPV, NPV, and F1 scores were calculated. 95% Confidence Intervals (95% CI) were calculated using profile likelihood. When values were 0% or 100%, exact (Clopper-Pearson) intervals were reported. We evaluated overall differences in accuracy across MLLMs using Cochran's Q test.

Pairwise comparisons using McNemar's tests between MLLMs were performed following a statistically significant Cochran's Q. To control for multiple comparisons across the ten possible model pairs, a Bonferroni correction was applied, with statistical significance defined as $p<0.005$. The flow of data is shown in **Figure 1**.

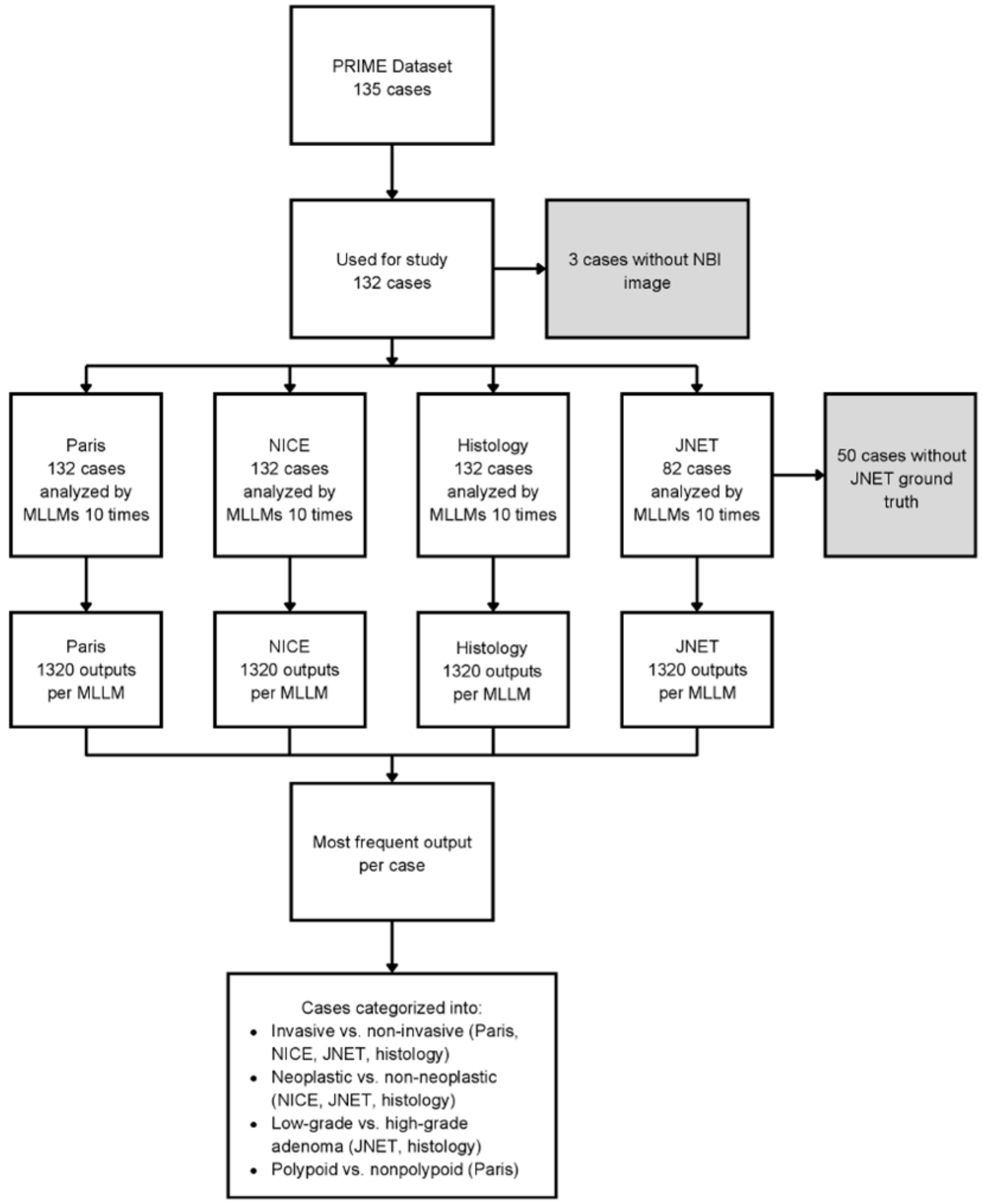

Figure 1: Flow diagram showing data handling.

MLLM outputs using the short prompt were also coded against the gold standard as “correct” or “incorrect”. Percent correct scores were calculated. A McNemar's test was used to compare long prompt outputs to the short prompt outputs. As five pair comparisons were conducted, statistical significance was set at $p<0.01$, following Bonferroni correction.

Statistical analyses were performed in SPSS (Version 31, IBM Corp., Armonk, NY). Crossstabulations and general linear model (GENLIN) methods were performed to generate sensitivity, specificity, PPV, NPV, accuracy, and 95% confidence interval estimates. MedCalc Software (Version 23.3.7) was used to calculate the exact (Clopper-Pearson) confidence intervals for 0% and 100% estimates [26]. Descriptive statistics, Cochran's Q, and McNemar's test values were calculated using SPSS. Statistical significance was defined as $p<0.05$ where no Bonferroni correction was applied.

## RESULTS

Of the 135 cases from the PRIME dataset, three were missing NBI images. Therefore, 132 were included for analyses of diagnostic accuracy for Paris, NICE, and predicted histology. For analyses on high grade adenoma vs. low grade adenoma, cases were omitted when the MLLM or the gold standard assigned a non-adenoma classification. A description of the PRIME dataset is shown in **Supplementary Table 1**.

### Paris

The percentage of correct responses ranged from 24.2% to 41.7% (**Table 1**). Cochran's Q was 32.9 ($p < 0.001$), indicating a statistically significant difference between models. Upon further analysis using pairwise McNemar's test, both Claude Opus 4 and GPT-5 demonstrated statistically significantly higher accuracy (41.7%), than Gemini 2.5 Pro, GPT-o3 and GPT-4o, with scores of 30.3%, 33.3%, and 24.2%, respectively. There was a statistically significant difference in the overall accuracies between the MLLMs when classifying polypoid vs. non-polypoid lesions, as indicated by the Cochran's Q (28.7, $p<0.001$) (**Supplementary Table 2**).

### NICE

The percentage of correct responses ranged from 78.0% to 84.1% (**Table 1**). Cochran's Q was 4.48 ($p=0.345$), indicating no statistically significant difference between the five MLLMs. F1 scores for invasive vs. non-invasive classifications ranged from 0.00 to 0.560 while the accuracy of the models ranged from 91.7% (95% CI: 85.6 to 95.8) to 94.7% (95% CI: 89.4 to 97.8) (**Table 2**). Cochran's Q was 2.51 ($p=0.642$), indicating no statistically significant difference between models. F1 scores for neoplastic vs. non-neoplastic classifications ranged from 0.910 to 0.937 while the accuracy of the models ranged from 84.1% (95% CI: 76.7 to 89.9) to 88.6% (95% CI: 82.0 to 93.5) (**Table 3**). Cochran's Q was 3.59 ($p=0.464$), indicating no statistically significant difference between models.

### Predicted Histology

The percentage of correct responses ranged from 55.3% to 62.9% (**Table 1**). Cochran's Q was 4.19 (p=0.381), indicating no statistically significant difference between the MLLMs. F1 scores for invasive vs. non-invasive classifications ranged from 0.00 to 0.500 while the accuracy of the models ranged from 92.4% (95% CI: 86.5 to 96.3) to 94.7% (95% CI: 89.4 to 97.8) (**Table 2**). Cochran's Q was 2.62 (p=0.624), indicating no statistically significant difference between models. F1 scores for neoplastic vs. non-neoplastic classifications ranged from 0.965 to 0.981 while the accuracy of the models ranged from 93.2% (95% CI: 87.4 to 96.8) to 96.2% (95% CI: 91.4 to 98.8) (**Table 3**). Cochran's Q was 6.08 (p=0.193), indicating no statistically significant difference between models. F1 scores for low-grade vs. high-grade classifications ranged from 0.00 to 0.492 while the accuracy of the models ranged from 67.0% (95% CI: 56.9 to 76.1) to 75.5% (95% CI: 66.0 to 83.5) (**Table 4**). Cochran's Q was 7.97 (p=0.093), which is suggestive of statistical difference between models.

### Short vs. Long Prompt Modalities

With NICE, Claude Opus 4 and Gemini 2.5 Pro achieved higher accuracy with the long prompt, at 84.8% and 78.8% for the long prompt vs 81.8% and 73.5% for the short prompt respectively, (both McNemar p<0.001), whereas GPT-o3/4o/5 performed better with the short prompt (all p<0.001). For Paris and Histology, several models showed statistically significant McNemar's test values, including cases with identical accuracies, as seen with Claude Opus 4 and GPT-4o for Paris, and GPT-o3 and -4o for Histology, indicating asymmetric disagreements between prompts (**Table 5**).

## DISCUSSION

This study investigated the accuracy of MLLMs in the optical diagnosis of colorectal polyps alongside characterization via the Paris and NICE classification systems. The results demonstrated that the five MLLMs in this study provide varying degrees of success. Models exceeded literature-reported trainee accuracy in the NICE classification method. We compared our results to ASGE and ESGE guidelines, the only existing guidelines from large societies which outline the sensitivity, specificity, and NPV for adenomatous histology that are recommended for effective optical diagnosis [3, 28]. For a diagnose-and-leave strategy, the ASGE guidelines recommend a NPV ≥ 90% [28]. The ESGE guidelines recommend a sensitivity ≥ 90% and a specificity ≥ 80% [3]. We also used human endoscopist and traditional CADx system performance to benchmark our results [27, 29]. MLLM performance remained below guideline thresholds despite high accuracy in some categories.

Models showed modest ability to diagnose invasive carcinoma using the NICE classification mode, with Gemini 2.5 Pro achieving the highest sensitivity at 70.0% (95% CI: 39.3 to 91.5) and an F1 score of 0.560. Other models struggled with classifying invasive carcinoma, as indicated by F1 scores of 0.00 or 0.222 for histology. This may be indicative of a tendency of MLLMs to bias towards more common classifications, in particular NICE Type 2. Prospectively, it could be valuable to explore whether systems such as magnifying endoscopy with NBI and blue light imaging (BLI) could mitigate MLLM biases and improve polyp characterization, particularly in differentiating low- and high-grade dysplasia [30].

Sensitivity, PPV, and overall accuracy remained high (>80%) for all models for neoplastic vs. non-neoplastic polyps. Interestingly, Bouwens et al. (2013) found that trainee endoscopists had an average accuracy of 84% in identifying adenomatous polyps using standard histology classification, which is slightly lower than our reported accuracy of MLLMs in identifying neoplastic polyps [27] (Table 2). Specificity and NPV were low, however, with Claude Opus 4 leading the others in all classification modes (e.g. 33.3% and 66.7% respectively in the NICE category). This may be reflective of a low incidence of hyperplastic polyps (n=5) in the dataset. Given current ESGE guidelines for the diagnose-and-leave strategy, it is clear that further research is required to determine the ability of MLLMs to reliably differentiate neoplastic from hyperplastic polyps [3].

When compared to a study conducted by Massimi et al. (2025), GPT-4o had a lower sensitivity, specificity, and accuracy in the differentiation between polypoid and non-polypoid lesions using the Paris classification [30] (**Supplementary Table 2**). The database they used includes videos of polyps, with five frames of each lesion captured at different angles sent to the MLLM for analysis. The PRIME dataset, however, contains only one perspective per polyp in WLE and NBI. Additionally, the SUN database is free to access while the PRIME dataset is paid. Therefore, it is possible that the SUN dataset was used in MLLM training. Further consideration must be placed into dataset novelty in MLLM studies as well as whether alternative views improve polyp classification.

Overall, Claude Opus 4 and Gemini 2.5 Pro were the two most promising models tested, with Claude leading in the differentiation of neoplastic vs. non-neoplastic polyps and Gemini leading in the differentiation of invasive vs. non-invasive and low vs. high-grade adenomatous polyps.

CADx models have an overall higher accuracy for histology than the MLLMs tested in this study. They, however, only produce a heat map and lack the ability to provide justification for diagnoses in terms that are easily understood by the endoscopist [29]. Furthermore, MLLMs can provide a cost-effective alternative to systems like CADe and may be capable of doing the same for CADx [19]. This warrants further research into novel, human-in-the-loop workflows that pair trainees with MLLMs, allowing them to supply basic information such as colonic segment and size to allow the MLLM to better characterize the polyp and provide trainees guidance to enhance their learning. Future studies should compare diagnostic accuracy and confidence for trainees alone, MLLMs alone, and trainees with MLLM assistance.

In our secondary analyses, we found that long prompts generally produce better diagnostic predictions than short prompts, as shown by Claude Opus 4 and Gemini 2.5 Pro using the NICE classification and predicted histology, as well as GPT-o3 and GPT-5 using the Paris classification. This prompting approach has been shown to improve fidelity to diagnostic criteria and allow for automated validation and analysis [16, 17, 23]. The long prompt modality does indeed show statistically significantly higher scores, particularly for GPT-5 in the Paris classification. The reverse findings, however, were found for the short prompting of GPT-o3/4o/5 for NICE classification, which goes against what literature has shown [17, 23]. For Paris and histology, the presence of statistically significant McNemar's test values despite identical accuracies, suggests that overall accuracy can obscure important asymmetries in model predictions. This highlights the importance of evaluating agreement and errors between individual cases, rather than accuracy alone, when benchmarking MLLMs for medical image classification.

These findings could be affected by whether the underlying MLLM was optimized for explicit reasoning. Reasoning-focused multimodal models (Claude Opus 4, Gemini 2.5 Pro, GPT-o3, GPT-5) allocate additional computation to structured multi-step reasoning, often using internal CoT processes, whereas generalist multimodal models such as GPT-4o emphasize speed and fluency at the expense of structured logic [31, 32]. This could explain why GPT-4o demonstrated the lowest percent correct score among the five MLLMs, as well as why reasoning MLLMs had higher accuracy than GPT-4o in differentiating polypoid vs. non-polypoid lesions. This does not, however, fully explain why its sensitivity, PPV and accuracy values are comparable when determining neoplasia, invasiveness and grade of adenoma. Further analyses will need to be conducted as to what effect the prompts and the models' inbuilt CoT reasoning have on diagnostic accuracy.

This analysis carries several limitations. The PRIME dataset is retrospective, which constrains generalizability. The dataset was not curated for a diagnostic accuracy study, and as such, certain polyp types, such as hyperplastic and Paris 0-III, lack sufficient representation. Additionally, because the dataset was curated for human evaluation, some images lack the quality ideal for MLLM diagnoses [33]. Images offer a single view of the polyp and lack size and location metadata. MLLM parameters, such as temperature and top p, may vary across providers.

## CONCLUSION

MLLMs show substantial promise in the optical diagnosis of colorectal polyps. These results justify prospective, multi center trials and careful human-in-the-loop design to consider their potential use for clinical practice and endoscopy training.

Table 1. Percent correct scores for all MLLMs in Paris, NICE, and histology.

| | Paris (n=132) | | NICE (n=132) | | Histology (n=132) | |
|---|---|---|---|---|---|---|
| MLLM | Correct (%) | Cochran's Q (p-value) | Correct (%) | Cochran's Q (p-value) | Correct (%) | Cochran's Q (p-value) |
| Claude Opus 4 | 41.7[a] | 32.9 (<0.001) | 84.1 | 4.48 (0.34) | 62.1 | 4.19 (0.38) |
| Gemini 2.5 Pro | 30.3[b] | | 78.8 | | 55.3 | |
| GPT-o3 | 33.3[c] | | 78.0 | | 60.6 | |
| GPT-4o | 24.2[d] | | 80.3 | | 60.6 | |
| GPT-5 | 41.7[e] | | 78.8 | | 62.9 | |

Pairwise comparisons showed statistical difference ($p<0.005$) based on McNemar's test (a) with Gemini 2.5 Pro, GPT-o3, and GPT-4o; (b) with Claude Opus 4 and Gemini 2.5 Pro; (c) with Claude Opus 4, GPT-4o, and GPT-5, (d) with Claude Opus 4, GPT-o3, and GPT-5; (e) with Gemini 2.5 Pro, GPT-o3, and GPT-4o

Table 2. Sensitivity, Specificity, PPV, NPV, Accuracy, and F1 scores of MLLMs for diagnosing non-invasive vs. invasive polyps.

| Invasive vs. Non-Invasive | | | | | | | | |
|---|---|---|---|---|---|---|---|---|
| Classification | MLLM | Sensitivity % (95% CI) | Specificity% (95% CI) | PPV % (95% CI) | NPV % (95% CI) | Accuracy % (95% CI) | F1 Score | Cochran's Q (p-value) |
| NICE (Type 3 vs. Rest) | Claude Opus 4 (n=132) | 30.0 (8.5 to 60.7) | 100 (97.0 to 100) | 100 (29.2 to 100) | 94.6 (89.8 to 97.6) | 94.7 (89.4 to 97.8) | 0.462 | 2.51 (0.64) |
| | Gemini 2.5 Pro (n=132) | 70.0 (39.3 to 91.5) | 93.4 (88.1 to 96.9) | 46.7 (23.5 to 70.9) | 97.4 (93.5 to 99.4) | 91.7 (85.6 to 95.8) | 0.560 | |
| | GPT-o3 (n=132) | 0.00 (0.00 to 30.9) | 100 (97.0 to 100) | N/A | 92.4 (87.1 to 96.1) | 92.4 (86.5 to 96.3) | 0.00 | |
| | GPT-4o (n=132) | 10.0 (0.60 to 37.2) | 100 (97.0 to 100) | 100 (2.50 to 100) | 93.1 (88.0 to 96.6) | 93.2 (87.4 to 96.8) | 0.182 | |
| | GPT-5 (n=132) | 10.0 (0.60 to 37.2) | 100 (97.0 to 100) | 100 (2.50 to 100) | 93.1 (91.7 to 94.3) | 93.2 (87.5 to 96.8) | 0.182 | |
| Histology ("Deep Invasive Carcinoma" vs. Rest) | Claude Opus 4 (n=132) | 12.5 (0.32 to 52.6) | 100 (97.1 to 100) | 100 (2.50 to 100) | 94.7 (93.2 to 95.8) | 94.7 (89.4 to 97.8) | 0.222 | 2.62 (0.62) |
| | Gemini 2.5 Pro (n=132) | 62.5 (24.5 to 91.5) | 94.4 (88.7 to 97.7) | 41.7 (22.5 to 63.7) | 97.5 (94.1 to 99.0) | 92.4 (86.5 to 96.3) | 0.500 | |
| | GPT-o3 (n=132) | 0.00 (0.00 to 36.9) | 100 (97.1 to 100) | N/A | 93.9 (93.9 to 93.9) | 93.9 (88.4 to 97.4) | 0.00 | |
| | GPT-4o (n=132) | 12.5 (0.32 to 52.6) | 100 (97.1 to 100) | 100 (2.50 to 100) | 94.7 (93.2 to 95.8) | 94.7 (89.4 to 97.8) | 0.222 | |
| | GPT-5 (n=132) | 12.5 (0.32 to 52.6) | 100 (97.1 to 100) | 100 (2.50 to 100) | 94.7 (93.2 to 95.8) | 94.7 (89.4 to 97.8) | 0.222 | |

CI, confidence interval

Table 3. Sensitivity, Specificity, PPV, NPV, Accuracy, and F1 scores of MLLMs for diagnosing non-neoplastic vs neoplastic polyps.

| Neoplastic vs. Non-Neoplastic | | | | | | | | |
|---|---|---|---|---|---|---|---|---|
| Classification | MLLM | Sensitivity % (95% CI) | Specificity % (95% CI) | PPV % (95% CI) | NPV % (95% CI) | Accuracy % (95% CI) | F1 Score | Cochran's Q (p-value) |
| NICE (Rest vs. Type 1) | Claude Opus 4 (n=132) | 97.4 (93.3 to 99.3) | 33.3 (14.8 to 56.3) | 90.2 (14.8 to 56.3) | 66.7 (34.5 to 90.5) | 88.6 (82.0 to 93.5) | 0.937 | 3.59 (0.46) |
| | Gemini 2.5 Pro (n=132) | 100.0 (96.8 to 100) | 0.00 (0.00 to 18.5) | 86.4 (79.8 to 91.5) | N/A | 86.4 (79.3 to 91.7) | 0.927 | |
| | GPT-o3 (n=132) | 93.0 (82.7 to 93.8) | 27.8 (15.9 to 65.2) | 89.1 (87.3 to 97.7) | 38.5 (11.0 to 50.5) | 84.1 (76.7 to 89.9) | 0.910 | |
| | GPT-4o (n=132) | 98.2 (94.7 to 99.7) | 11.1 (1.9 to 30.5) | 87.5 (81.0 to 92.5) | 50.0 (10.7 to 89.3) | 86.4 (79.3 to 91.7) | 0.926 | |
| | GPT-5 (n=132) | 93.0 (87.3 to 96.7) | 33.3 (14.8 to 56.3) | 89.8 (83.5 to 94.4) | 43.9 (19.8 to 68.3) | 84.8 (77.6 to 90.5) | 0.914 | |
| Histology (Rest vs. "Hyperplastic polyp") | Claude Opus 4 (n=132) | 95.3 (90.0 to 98.2) | 60.0 (14.7 to 94.7 | 98.4 (95.4 to 99.4) | 33.3 (14.8 to 59.0) | 93.9 (88.4 to 97.4) | 0.968 | 6.08 (0.19) |
| | Gemini 2.5 Pro (n=132) | 100 (97.1 to 100) | 0.00 (0.00 to 52.2) | 96.2 (96.2 to 96.2) | N/A | 96.2 (91.4 to 98.8) | 0.981 | |
| | GPT-o3 (n=132) | 100 (97.1 to 100) | 0.00 (0.00 to 52.2) | 96.2 (96.2 to 96.2) | N/A | 96.2 (91.4 to 98.8) | 0.981 | |
| | GPT-4o (n=132) | 96.9 (92.1 to 99.1) | 0.00 (0.00 to 52.2) | 96.1 (96.0 to 96.2) | N/A | 93.2 (87.4 to 96.8) | 0.965 | |
| | GPT-5 (n=132) | 100 (97.1 to 100) | 0.00 (0.00 to 52.2) | 96.2 (96.2 to 96.2) | N/A | 96.1 (91.2 to 98.7) | 0.981 | |

CI, confidence interval

Table 4. Sensitivity, Specificity, PPV, NPV, Accuracy, and F1 scores of MLLMs for diagnosing low- vs. high-grade dysplasia.

| Low-Grade vs. High-Grade Adenoma | | | | | | | | |
|---|---|---|---|---|---|---|---|---|
| Classification | MLLM | Sensitivity % (95% CI) | Specificity % (95% CI) | PPV % (95% CI) | NPV % (95% CI) | Accuracy % (95% CI) | F1 Score | Cochran's Q (p-value) |
| Histology ("Adenoma" vs. "High Risk Adenoma or Superficial Adenocarcinoma") | Claude Opus 4 (n=105) | 3.70 (0.09 to 19.0) | 100 (95.4 to 100) | 100 (2.50 to 100) | 75.0 (73.6 to 76.4) | 75.2 (65.9 to 83.1) | 0.071 | 7.97 (0.09) |
| | Gemini 2.5 Pro (n=100) | 66.7 (44.7 to 84.4) | 67.1 (55.4 to 77.5) | 39.0 (29.4 to 49.5) | 86.4 (78.0 to 92.0) | 67.0 (56.9 to 76.1) | 0.492 | |
| | GPT-o3 (n=100) | 0.00 (0.00 to 13.7) | 100 (95.2 to 100) | N/A | 75.0 (75.0 to 75.0) | 75.0 (65.3 to 83.1) | 0.00 | |
| | GPT-4o (n=106) | 0.00 (0.00 to 12.3) | 100 (95.4 to 100) | N/A | 73.6 (73.6 to 73.6) | 73.6 (64.1 to 81.7) | 0.00 | |
| | GPT-5 (n=102) | 3.80 (0.10 to 19.6) | 100 (95.3 to 100) | 100 (2.50 to 100) | 75.2 (73.8 to 76.6) | 75.5 (66.0 to 83.5) | 0.074 | |

CI, confidence interval

Table 5. Percent correct scores of the MLLMs for Paris, NICE, and Histology classifications for both long and short prompt modalities.

| | Paris (n=132) | | | NICE (n=132) | | | Histology (n=132) | | |
|---|---|---|---|---|---|---|---|---|---|
| MLLM | Long (%) | Short (%) | McNemar's Test p-value | Long (%) | Short (%) | McNemar's Test p-value | Long (%) | Short (%) | McNemar's Test p-value |
| Claude Opus 4 | 41.7 | 41.7 | 0.07 | 84.1 | 81.8 | <0.001* | 62.1 | 56.8 | 0.03 |
| Gemini 2.5 pro | 30.3 | 41.7 | 0.07 | 78.8 | 73.5 | <0.001* | 55.3 | 50.8 | 0.53 |
| GPT-o3 | 33.3 | 30.3 | <0.001* | 78.0 | 79.5 | <0.001* | 60.6 | 60.6 | 0.02 |
| GPT-4o | 24.2 | 24.2 | <0.001* | 80.3 | 81.1 | <0.001* | 60.6 | 60.6 | 0.02 |
| GPT-5 | 41.7 | 31.8 | <0.002* | 78.8 | 81.1 | <0.001* | 62.9 | 60.6 | 0.02 |

McNemar's test compared discordant pairs. A significant p-value can occur even when accuracies are equal if the prompts are correct on different subsets of items. (*) p-value <0.01, indicating statistically significant difference between short and long prompt modalities.

## REFERENCES

1. Siegel RL, Wagle NS, Cercek A, et al. Colorectal cancer statistics, 2023. CA Cancer J Clin. 2023 May-Jun;73(3):233-254. doi: 10.3322/caac.21772. Epub 2023 Mar 1.
2. Nagarajan KV, Bhat N. Imaging colonic polyps in 2024. Indian J Gastroenterol. 2024 Oct;43(5):954-965. doi: 10.1007/s12664-024-01679-y. Epub 2024 Sep 30.
3. Ferlitsch M, Hassan C, Bisschops R, et al. Colorectal polypectomy and endoscopic mucosal resection: European Society of Gastrointestinal Endoscopy (ESGE) Guideline - Update 2024. Endoscopy. 2024 Jul;56(7):516-545. doi: 10.1055/a-2304-3219. Epub 2024 Apr 26.
4. Participants in the Paris Workshop. The Paris endoscopic classification of superficial neoplastic lesions: esophagus, stomach, and colon. Gastrointestinal Endoscopy. 2003 Dec;58(6):S3–43.
5. Hayashi N, Tanaka S, Hewett DG, et al. Endoscopic prediction of deep submucosal invasive carcinoma: validation of the narrow-band imaging international colorectal endoscopic (NICE) classification. Gastrointest Endosc. 2013 Oct;78(4):625-32. doi: 10.1016/j.gie.2013.04.185. Epub 2013 Jul 30.
6. Vleugels JLA, Koens L, Dijkgraaf MGW, et al. Suboptimal endoscopic cancer recognition in colorectal lesions in a national bowel screening programme. Gut. 2020 Jun;69(6):977-980. doi: 10.1136/gutjnl-2018-316882. Epub 2019 Dec 10. PMID: 31822579; PMCID: PMC7282551.
7. Hamada Y, Tanaka K, Katsurahara M, et al. Utility of the narrow-band imaging international colorectal endoscopic classification for optical diagnosis of colorectal polyp histology in clinical practice: a retrospective study. BMC Gastroenterol. 2021 Aug 28;21(1):336. doi: 10.1186/s12876-021-01898-z.
8. Kandel P, Wallace MB. Should We Resect and Discard Low Risk Diminutive Colon Polyps. Clin Endosc. 2019 May;52(3):239-246. doi: 10.5946/ce.2018.136. Epub 2019 Jan 21.
9. Pecere S, Antonelli G, Dinis-Ribeiro M, et al. Endoscopists performance in optical diagnosis of colorectal polyps in artificial intelligence studies. United European Gastroenterol J. 2022 Oct;10(8):817-826. doi: 10.1002/ueg2.12285. Epub 2022 Aug 19.
10. Joseph J, LePage EM, Cheney CP, Pawa R. Artificial intelligence in colonoscopy. World J Gastroenterol. 2021 Aug 7;27(29):4802-4817. doi: 10.3748/wjg.v27.i29.4802.
11. Spadaccini M, Iannone A, Maselli R, et al. Computer-aided detection versus advanced imaging for detection of colorectal neoplasia: a systematic review and network meta-

analysis. Lancet Gastroenterol Hepatol. 2021 Oct;6(10):793-802. doi: 10.1016/S2468-1253(21)00215-6.

12. Deng J, Heybati K, Shammas-Toma M. When vision meets reality: Exploring the clinical applicability of GPT-4 with vision. Clin Imaging. 2024 Apr;108:110101. doi: 10.1016/j.clinimag.2024.110101.
13. Huppertz MS, Siepmann R, Topp D, et al. Revolution or risk?-Assessing the potential and challenges of GPT-4V in radiologic image interpretation. Eur Radiol. 2025 Mar;35(3):1111-1121. doi: 10.1007/s00330-024-11115-6. Epub 2024 Oct 18.
14. Tomita K, Nishida T, Kitaguchi Y, et al. Image Recognition Performance of GPT-4V(ision) and GPT-4o in Ophthalmology: Use of Images in Clinical Questions. Clin Ophthalmol. 2025 May 8;19:1557-1564. doi: 10.2147/OPTH.S494480.
15. Ferber D, Wölflein G, Wiest IC, et al. In-context learning enables multimodal large language models to classify cancer pathology images. Nat Commun. 2024 Nov 21;15(1):10104. doi: 10.1038/s41467-024-51465-9.
16. OpenAI. Thinking with images. 2025.
17. Bhayana R. Chatbots and Large Language Models in Radiology: A Practical Primer for Clinical and Research Applications. Radiology. 2024 Jan;310(1):e232756. doi: 10.1148/radiol.232756.
18. Carlini L, Massimi D, Mori Y, et al. Large language models for detecting colorectal polyps in endoscopic images. Gut. 2025 May 24:gutjnl-2025-335091. doi: 10.1136/gutjnl-2025-335091. Epub ahead of print. PMID: 40360230.
19. Walsh CM, Grover SC, Khan R, et al. ImageSIM PRIME, adult gastrointestinal polyp – diagnosis & management [Internet]. Available from: https://imagesimcme.com/course/adult-gastrointestinal-polyp-diagnosis-and-management.
20. Cohen JF, Korevaar DA, Altman DG, et al. STARD 2015 guidelines for reporting diagnostic accuracy studies: explanation and elaboration. BMJ Open. 2016 Nov 14;6(11):e012799. doi: 10.1136/bmjopen-2016-012799. PMID: 28137831; PMCID: PMC5128957.
21. White C, Dooley S, Roberts M, et al. LiveBench: A challenging, contamination-free LLM benchmark. In: Proceedings of the Thirteenth International Conference on Learning Representations; 2025.
22. Close encounters of the AI kind - Imagining the Digital Future Center [Internet]. Imagining the Digital Future Center. 2025. Available from:

https://imaginingthedigitalfuture.org/reports-and-publications/close-encounters-of-the-ai-kind/

23. Kusano G, Akimoto K, Takeoka K. Are Longer Prompts Always Better? Prompt Selection in Large Language Models for Recommendation Systems. arXiv.org. Published 2024. https://arxiv.org/abs/2412.14454
24. Liu Q, Wang W, Willard J. Effects of Prompt Length on Domain-specific Tasks for Large Language Models. arXiv.org. Published 2025. http://arxiv.org/abs/2502.14255
25. Fujiyoshi MRA, Fujiyoshi Y, Gimpaya N, et al. Unified Magnifying Endoscopic Classification (UMEC) of Gastrointestinal Lesions: A North American Validation Study. J Can Assoc Gastroenterol. 2023 Dec 9;7(3):246-254. doi: 10.1093/jcag/gwad055. PMID: 38841140; PMCID: PMC11149659.
26. MedCalc Software Ltd. Diagnostic test evaluation calculator. https://www.medcalc.org/en/calc/diagnostic_test.php (Version 23.3.7)
27. Bouwens MW, de Ridder R, Masclee AA, et al. Optical diagnosis of colorectal polyps using high-definition i-scan: an educational experience. World J Gastroenterol. 2013 Jul 21;19(27):4334-43. doi: 10.3748/wjg.v19.i27.4334. PMID: 23885144; PMCID: PMC3718901.
28. Parsa N, Rex DK, Byrne MF. Colorectal polyp characterization with standard endoscopy: Will Artificial Intelligence succeed where human eyes failed? Best Pract Res Clin Gastroenterol. 2021 Jun-Aug;52-53:101736. doi: 10.1016/j.bpg.2021.101736. Epub 2021 Feb 22. PMID: 34172255.
29. Nazarian S, Glover B, Ashrafian H, et al. Diagnostic Accuracy of Artificial Intelligence and Computer-Aided Diagnosis for the Detection and Characterization of Colorectal Polyps: Systematic Review and Meta-analysis. J Med Internet Res. 2021 Jul 14;23(7):e27370. doi: 10.2196/27370. PMID: 34259645; PMCID: PMC8319784.
30. Massimi D, Carlini L, Mori Y, et al. Large language model for interpreting the Paris classification of colorectal polyps. Endosc Int Open. 2025 Oct 9;13:a27030209. doi: 10.1055/a-2703-0209. PMID: 41079216; PMCID: PMC12511921.
31. Li ZZ, Zhang D, Zhang ML, et al. From System 1 to System 2: A Survey of Reasoning Large Language Models. arXiv [csAI]. Published online 2025. http://arxiv.org/abs/2502.17419
32. Xu P, Wu Y, Jin K, et al. DeepSeek-R1 outperforms Gemini 2.0 Pro, OpenAI o1, and o3-mini in bilingual complex ophthalmology reasoning. Advances in Ophthalmology Practice and Research. 2025;5(3):189–95. doi: 10.1016/j.aopr.2025.05.001.

33. Cheng Z, Ong A, Wagner S, et al. Understanding the robustness of vision-language models to medical image artefacts. MedRxiv. Published 2025. https://www.medrxiv.org/content/10.1101/2025.05.13.25327495v1